\documentclass[11pt]{article}

\usepackage[preprint]{acl}
\usepackage{times}
\usepackage{latexsym}
\usepackage[T1]{fontenc}
\usepackage[utf8]{inputenc}
\usepackage{microtype}
\usepackage{flafter}
\usepackage{inconsolata}

\usepackage{booktabs}
\usepackage{amsfonts}
\usepackage{amsmath}
\usepackage{nicefrac}
\usepackage{graphicx}
\usepackage{tikz}
\usetikzlibrary{shadows,shapes,arrows,positioning}
\usepackage{pgfplots}
\pgfplotsset{compat=1.18}
\usepackage{listings}
\usepackage{float}

\title{Losing Sight of the Image in Long Chart Extraction}

\author{%
  Shrish Goel \\
  California Institute \\
  of Technology \\
  \texttt{sgoel2@caltech.edu} \\
  \And
  Tridip Das \\
  FAMU-FSU College \\
  of Engineering \\
  \texttt{tdas@eng.famu.fsu.edu} \\
  \And
  Dan Takahashi \\
  Takahuman \\
  \texttt{dan@takahuman.com} \\
}

\begin{document}

\maketitle
\begin{abstract}
Vision-language models (VLMs) are increasingly used to turn scientific figures into structured data, a task that requires long outputs grounded in a single image. Chart digitization makes this grounding measurable, because every emitted coordinate has a verifiable referent in the image. We evaluate five VLMs from four model families, including a Gemma~4~31B fine-tuned on 22{,}285 annotated biomedical charts, on 175 real figure panels and 300 synthetic charts with exhaustive ground truth. On dense scatter plots, the share of extracted points that match a real marker is lower late in the output than early for every model, and the decline persists after controlling for incomplete annotation. Two interventions provide evidence that emission order, not only chart region, contributes to this loss. Reversing the emission order changes where precision falls in GPT-5.6 Luna, and mirroring synthetic charts, which moves the same content from the end of the output to the start, lowers the precision of content written late in all five models. Longer outputs also fail qualitatively, as models copy earlier series, invent regular grids, and emit points inside regions that are verifiably empty. Splitting charts into tiles raises recall from 48\% to 67\%, yet precision decays again within tiles, suggesting that accuracy depends largely on how much the model has generated rather than how much of the image it must cover. Because these errors pass schema validation, they are silent in downstream pipelines.
\end{abstract}
\section{Introduction}
\label{sec:intro}

Vision-language models (VLMs) are increasingly asked to produce long, structured outputs from a single image. They transcribe tables, convert documents into records, and digitize figures into machine-readable data. Research on long contexts has focused mainly on the input side, asking how well models retrieve and reason over extended prompts \citep{liu_lost_2023}. The output side has received far less attention. When a model must write thousands of tokens conditioned on one fixed image, it is unclear whether each later token remains as faithful to the image as the first. Evidence from multimodal reasoning suggests it may not, since attention to image tokens tends to decline as generation proceeds \citep{chung_v1_2025}. Most such evidence, however, comes from tasks where outputs are short and individual tokens cannot be checked against the image.

Scientific chart digitization removes both limitations. A large share of published experimental data survives only as plotted figures, and recovering it at scale requires automated extraction rather than manual tools such as WebPlotDigitizer \citep{rohatgi_webplotdigitizer}. A dense scatter plot requires a model to emit hundreds of coordinates in one pass, and every coordinate has a referent in the image that can be verified. Chart digitization therefore offers a long output horizon together with dense, per-token supervision, which makes it an unusually precise instrument for asking whether visual grounding survives long generation. The answer also matters in practice. Extraction errors that look like plausible data would enter literature-mining databases unnoticed.

We study this question across five VLMs from four model families. They include GPT-5.6 Luna, Claude Haiku 4.5, Gemma~4~31B, Qwen3.8 Flash, and a Gemma~4~31B fine-tuned on 22{,}285 annotated biomedical charts. We evaluate them on 175 panels from real biomedical figures and on 300 synthetic charts whose ground truth is exhaustive by construction. Our central measurement is point precision at matched output positions, the share of extracted points that lie near a real marker, computed within each chart for early versus late portions of the output. Because real annotations subsample dense point clouds, we pair this measurement with a split-half null that estimates how much precision would fall from annotation sparsity alone.

The results show a consistent pattern. On dense scatter plots, precision in the last quarter of the output is lower than in the first for every model, by 10 to 22 percentage points on real charts. The decline exceeds the annotation-sparsity null by 5 to 14 points and remains significant for every model on synthetic charts with complete ground truth. Because models write points from left to right, output position and horizontal position on the chart are confounded. Reversing the emission order in GPT-5.6 Luna reduces the excess decline toward the right side of the chart, and mirroring synthetic charts, which moves the same content from the end of the output to the start, shows that, on single-series charts, the same content is 5.8 to 23.9 points less precise when written late in all five models.

Long outputs also fail abruptly. Models abandon the image and write copied series, invented regular grids, and arithmetic progressions, a failure we call collapse, and every model writes points inside chart regions that are verifiably empty. Fine-tuning does not prevent these failures. Splitting charts into tiles raises recall from 48\% to 67\%, but precision falls again within tile outputs, so accuracy depends largely on how much the model has already written, not on how large a region of the image it must read.

Our contributions are as follows.
\begin{itemize}
    \item We introduce chart digitization as a testbed for output-side grounding, with position-matched precision, an annotation-sparsity null, and a synthetic benchmark with exhaustive ground truth (Section~\ref{sec:setup}).
    \item We show that precision declines with output position across five VLMs from four families, beyond what incomplete annotation explains (Section~\ref{sec:decline}).
    \item We provide evidence from two interventions, prompt reversal and chart mirroring, that emission order contributes to the decline in all five models (Section~\ref{sec:causal}).
    \item We characterize collapse and fabrication in long outputs, including points written into verifiably empty regions, and show that forcing longer outputs makes collapse more common (Section~\ref{sec:collapse}).
    \item We show that chunked extraction improves completeness but not faithfulness, because precision decays again within chunks, and we relate collapse to reduced attention to the image (Sections~\ref{sec:chunking} and~\ref{sec:attention}).
\end{itemize}

\section{Related Work}
\label{sec:related}

\paragraph{Chart understanding and extraction.}
Benchmarks such as PlotQA \citep{methani_plotqa_2020}, ChartQA \citep{masry_chartqa_2022}, and CharXiv \citep{wang_charxiv_2024} evaluate charts mainly through question answering, where answers are short and a model can succeed without transcribing every data point. Chart-to-table models such as DePlot \citep{liu_deplot_2023} and MatCha \citep{liu_matcha_2023} target full data recovery but are trained largely on clean plots with flat table targets, and they degrade on richer scientific structures. PlotPick, for example, finds DePlot competitive on standard chart types but far behind general-purpose VLMs on box plots \citep{carstensen_plotpick_2026}. ChartLlama \citep{han_chartllama_2023}, StructChart \citep{xia_structchart_2023}, and ChartX \citep{xia_chartx_2025} broaden chart understanding toward structured extraction and reasoning, and ChartRecover reconstructs data through element detection and tick-based coordinate transforms \citep{yuan_deep_2026}. These works report extraction quality as an aggregate over each chart. We instead treat extraction as a long-generation problem and measure accuracy as a function of where each point appears in the output, which aggregate metrics cannot reveal.

\paragraph{Hallucination in vision-language models.}
Object hallucination, where models describe content absent from the image, was first quantified in image captioning \citep{rohrbach_object_2018} and later benchmarked for large VLMs \citep{li_evaluating_2023}. Several studies link hallucination to position and length. Objects mentioned later in a caption are more often hallucinated \citep{zhou_analyzing_2024}, and hallucination grows with output length as the influence of visual input fades \citep{favero_multimodal_2024}. Mechanistic accounts point to the same shift. Models attend little to image tokens in deeper layers \citep{chen_image_2024}, over-trust a few previously generated tokens \citep{huang_opera_2024}, and lose attention to the image as reasoning chains lengthen \citep{chung_v1_2025}. Proposed remedies operate at decoding time, for instance by contrasting outputs with and without the image \citep{leng_mitigating_2024, favero_multimodal_2024}. Much of this work evaluates captions, object judgments, or reasoning chains, where individual tokens are hard to verify against the image. Chart extraction offers dense point-level correctness, which lets us measure decline by output position, check it against annotation sparsity, and test the role of emission order with two ordering interventions.

\paragraph{Long sequences and degenerate generation.}
On the input side, models retrieve information less reliably from the middle of long contexts \citep{liu_lost_2023}, concentrate attention on initial sink tokens \citep{xiao_efficient_2023}, and lose accuracy when irrelevant images are added to the input \citep{sharma_losing_2024}. On the output side, open-ended generation is prone to repetition and degenerate loops \citep{holtzman_curious_2020, welleck_neural_2020, xu_learning_2022}, often attributed to exposure bias, the mismatch between training on reference prefixes and generating from a model's own prefix \citep{bengio_scheduled_2015, ranzato_sequence_2016}. The collapse we observe in chart extraction is related to degeneration, yet it takes a form that standard repetition checks miss. Models write invented grids and arithmetic progressions that are valid, non-repeating, and plausible, and they pass schema validation because the format remains intact while the content stops tracking the image.

\section{Experimental Setup}
\label{sec:setup}

\subsection{Models}
\label{sec:models}

We evaluate five VLMs from four model families. Four are used zero-shot, namely GPT-5.6 Luna, Claude Haiku 4.5, Gemma~4~31B, and Qwen3.8 Flash, queried through OpenRouter with greedy decoding and a limit of 32{,}000 output tokens, with thinking disabled for Qwen3.8 Flash. The fifth is a Gemma~4~31B fine-tuned with QLoRA on 22{,}285 charts from 3{,}628 PubMed Central figures, with a document-level split separating training and test articles (Appendix~\ref{app:pipeline}). It is served with vLLM in fp8 with an 8{,}000-token limit and its own training prompt, so comparisons with it reflect model and setup together. All models receive the full figure and a crop of the target panel, preprocessed identically.

\subsection{Data}
\label{sec:data}

\paragraph{Real charts.}
We draw 175 panels, 115 scatter plots and 60 line charts, from held-out biomedical figures in PubMed Central. Selection prioritizes dense panels with at least 100 annotated points, falls back to panels with 50 to 99 points, and takes at most two panels per article, yielding 175 panels from 162 articles. Ground truth consists of human-verified annotations of axes, legends, and data points. For dense scatter plots, annotators record a representative subset of points rather than every marker, and the annotation-sparsity null in Section~\ref{sec:metrics} accounts for this when comparing early and late portions of the output.

\paragraph{Synthetic charts.}
To obtain exhaustive ground truth, we generate 300 charts styled after biomedical figures (Appendix~\ref{app:synthetic}). The 200 scatter plots span three density tiers (50 to 100, 100 to 250, and 250 to 600 points), one to eight series, and five layouts, and the 100 line charts contain two to six series. Axis scales, styles, and multi-panel embedding are randomized. Every plotted point is recorded, including markers hidden by others, and in 60 scatter plots a rectangular region is deliberately left empty, so points written there have no marker beneath them, apart from rare cases near the boundary that fall within the matching tolerance of a marker outside. For the mirroring test in Section~\ref{sec:causal}, we also render a mirrored copy of each of the 200 scatter plots and generate 100 additional single-series scatter plots, each with a mirrored copy, balanced across density tiers and layouts (Appendix~\ref{app:mirror}).

\subsection{Prompting Conditions}
\label{sec:prompts}

The baseline prompt (B) asks the model to digitize every visible marker into the YAML schema, without instructions to subsample. The extended prompt (E) adds one statement, that the panel contains at least $N$ visible points and the output must contain at least $N$, where $N$ is the annotated point count, which encourages longer outputs without changing the task. On synthetic charts, $N$ includes markers hidden behind others. The reversed prompt (R) asks for points from right to left instead of in ascending order of $x$, leaving the rest of B unchanged.

\subsection{Metrics}
\label{sec:metrics}

\paragraph{Point precision and recall.}
We normalize both axes to the ground-truth range, using $\log_{10}$ on logarithmic axes. A predicted point is correct if a ground-truth point lies within distance $d=0.02$ of it, and precision is the fraction of predicted points that are correct. Recall is the fraction of ground-truth points with a predicted point within the same distance. Matching uses nearest neighbors without one-to-one assignment and ignores series labels. Duplicated predictions are not penalized as such, but they remain in the denominator and can shift precision. Results at $d=0.01$ and $d=0.05$, under one-to-one matching, with hidden markers removed, and with one panel per article are in Appendix~\ref{app:tables}.

\paragraph{Position in the output.}
We measure how precision changes along each output in two ways. Writing quarters split a chart's predicted points into four equal-count segments in emission order. Absolute bins group points by their index in the output (0 to 49, 50 to 99, 100 to 199, 200 to 399, and 400 or more), which compares models at the same distance into their outputs. All comparisons are paired within charts, so differences in difficulty between charts cannot produce them.

\paragraph{Annotation-sparsity null.}
On real charts, a correct point late in the output may be scored as wrong simply because annotators marked fewer points in that region. To estimate this effect, we split each chart's annotated points at random into halves $A$ and $B$, treat $A$ as a perfect prediction, sort it by $x$ into quarters, and score each quarter against $B$. Model predictions are scored against the same half $B$. We average over 20 random splits and report the excess decline, the model's first-to-last quarter change minus that of the perfect prediction.

\paragraph{Collapse.}
A rule-based detector flags four structural failure patterns, namely repeated blocks of points, series duplicated from another series, evenly spaced columns of points at a single $x$ value, in linear or logarithmic space, and arithmetic progressions, and records where in the output each pattern begins (Appendix~\ref{app:detector}). None of the 300 exact synthetic ground-truth point sets triggers the detector. A flag marks structural degeneration rather than a confirmed error, since approximate but correct coordinates can occasionally look regular.

\paragraph{Statistics.}
Charts are the unit of analysis. We report two-sided Wilcoxon signed-rank tests on paired within-chart differences, with unadjusted p-values. All results in Table~\ref{tab:decline}, the significant results in Table~\ref{tab:reversal}, the significant order effects in Table~\ref{tab:mirror}, and the robustness checks and mirrored-chart regressions in the appendix remain significant after Holm correction within each table. Precision and recall use complete outputs, while the annotation-sparsity and collapse analyses also include outputs recovered from truncated or partly malformed responses, so their denominators differ.
\section{Precision Declines with Output Position}
\label{sec:decline}

\begin{figure}[t]
\centering
\includegraphics[width=\columnwidth]{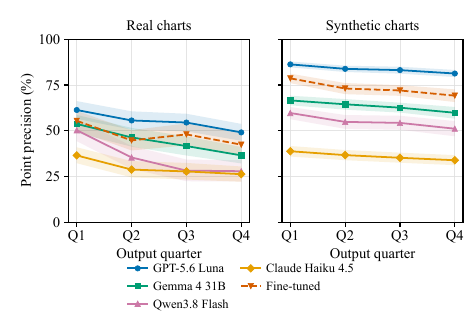}
\caption{Point precision on dense scatter plots by quarter of the output, with zero-shot models under the baseline prompt and the fine-tuned model under its own prompt. Precision falls from the first to the last quarter for every model, on real charts with human annotations (left) and on synthetic charts with exhaustive ground truth (right). Shading shows 95\% bootstrap confidence intervals over charts.}
\label{fig:quarters}
\end{figure}

\begin{table}[!htb]
\centering
\small
\setlength{\tabcolsep}{3.5pt}
\begin{tabular}{llccc}
\toprule
& & \multicolumn{2}{c}{\textbf{Real}} & \textbf{Synthetic} \\
\cmidrule(lr){3-4}\cmidrule(l){5-5}
\textbf{Model} & & $\Delta$ & Excess & $\Delta$ \\
\midrule
GPT-5.6 Luna & B & $-12.2$ & $-5.0$ & $-5.0$ \\
             & E & $-17.5$ & $-8.2$ & $-6.3$ \\
Claude Haiku 4.5 & B & $-10.2$ & $-7.9$ & $-4.9$ \\
                 & E & $-11.3$ & $-8.2$ & $-5.9$ \\
Gemma 4 31B & B & $-17.2$ & $-12.3$ & $-6.7$ \\
            & E & $-16.3$ & $-12.4$ & $-6.8$ \\
Qwen3.8 Flash & B & $-22.4$ & $-14.2$ & $-8.6$ \\
              & E & $-19.3$ & $-13.4$ & $-9.5$ \\
Fine-tuned & & $-13.0$ & $-9.1$ & $-9.4$ \\
\bottomrule
\end{tabular}
\caption{Precision change (percentage points) on scatter plots. $\Delta$ is the change from the first to the last quarter of the output in emission order, scored against all ground-truth points. Excess is a separate measurement, computed on quarters sorted by $x$ and scored against a held-out half of the annotations over 20 random splits, minus the decline of real annotated points scored the same way ($-4.3$ points for most models and $-4.1$ and $-3.5$ for Qwen3.8 Flash under B and E, whose contributing panels differ). All values are significant (real $\Delta$ $p<10^{-4}$, excess $p<0.01$, synthetic $\Delta$ $p<0.001$, Wilcoxon signed-rank).}
\label{tab:decline}
\end{table}

\paragraph{Every model loses precision as it writes.}
On real scatter plots, precision falls from the first to the last quarter of the output for every model and prompt (Figure~\ref{fig:quarters}, Table~\ref{tab:decline}). Under the baseline prompt, GPT-5.6 Luna drops from 61.4\% to 49.1\% and Qwen3.8 Flash from 50.3\% to 27.9\%. Grouping points by their index in the output gives the same result, with points 100 to 199 less precise than points 0 to 49 for every model, by 10.5 points for GPT-5.6 Luna and 16.8 points for the fine-tuned model ($p<0.001$ for both). Pairing within charts removes differences between charts. The decline also holds in synthetic outputs that contain no more points than the chart, and output position still predicts lower correctness after adjusting for location and marker density in 13 of 18 fits, with Claude Haiku 4.5 the main exception (Appendix~\ref{app:controls}).

\paragraph{The decline exceeds what incomplete annotation explains.}
Annotation density alone can produce a decline toward the right side of a chart. Real annotated points scored against a held-out half of the annotations lose 4.3 points between the first and last quarters sorted by $x$. Every model declines by more, with excess declines from 5.0 points for GPT-5.6 Luna to 14.2 points for Qwen3.8 Flash, all significant (Table~\ref{tab:decline}).

\paragraph{The decline persists with exhaustive ground truth.}
On synthetic charts, where every point is annotated, precision still falls from the first to the last quarter for every model and prompt, by 4.9 to 9.5 points ($p<0.001$). The declines are smaller than on real charts, consistent with a contribution from annotation density.

\paragraph{The decline is robust to scoring choices.}
The decline remains significant for every model and prompt under one-to-one matching, in which each ground-truth point can be matched at most once, when markers hidden behind others are removed from the synthetic ground truth, and when only one panel per article is kept (Appendix~\ref{app:tables}). Precision in the last quarter is also lower than in the first at tolerances of 0.01 and 0.05 in all 36 comparisons, significantly so in 31.

\paragraph{Fine-tuning does not prevent the decline.}
We detect no significant difference between the fine-tuned model and GPT-5.6 Luna on the first 50 points of real charts ($-2.1$ points, $p=0.58$), but the fine-tuned model is 8.2 points less precise on points 50 to 99 ($p=0.03$) and 16.7 points less precise on points 100 to 199 ($p<0.001$), on the charts where both models reach each position.

\section{Emission Order Contributes to the Decline}
\label{sec:causal}

Models write each series from left to right, so the right side of a chart tends to be written late, and the decline could reflect either a point's position in the output or harder reading on the right side, where data often thin out. Reversing the emission order separates the two. If the right side is simply harder, it should stay less precise when written first. If output position matters, whichever side is written last should lose precision.

\paragraph{Design.}
The reversed prompt R differs from B only in asking for points from right to left. In a five-chart compliance check, Claude Haiku 4.5 reversed no outputs, and Gemma~4~31B and Qwen3.8 Flash reversed line charts but not most scatter plots. Reversal would also be out of distribution for the fine-tuned model, which was trained on left-to-right targets. GPT-5.6 Luna complied, reversing a median of 98.6\% of consecutive point pairs on scatter plots, so we run R on all 175 real panels for GPT-5.6 Luna.

\begin{table}[t]
\centering
\small
\setlength{\tabcolsep}{3.5pt}
\begin{tabular}{lccccc}
\toprule
& \multicolumn{2}{c}{\textbf{Precision (\%)}} & \multicolumn{3}{c}{\textbf{Excess decline}} \\
\cmidrule(lr){2-3}\cmidrule(l){4-6}
\textbf{Order} & \textbf{Left} & \textbf{Right} & \textbf{All} & \textbf{$\geq$80\%} & \textbf{Single} \\
& & & (113) & (78) & (37) \\
\midrule
B (left first) & 60.9 & 48.5 & $-5.1^{**}$ & $-5.3^{**}$ & $-9.2^{**}$ \\
R (right first) & 58.2 & 51.7 & $-0.8$ & $-0.2$ & $+0.6$ \\
\bottomrule
\end{tabular}
\caption{GPT-5.6 Luna on real scatter plots, writing points left to right (B) or right to left (R). Left and right give precision on the first and last quarters of points sorted by $x$, scored against all annotations (113 paired charts). Excess decline is the left-to-right drop beyond that of real annotated points, scored against held-out halves of the annotations, for all charts, charts with at least 80\% of point pairs reversed, and single-series charts among those. $^{**}p<0.01$.}
\label{tab:reversal}
\end{table}

\paragraph{Reversal reduces the excess decline.}
Reversing the order reduces the excess left-to-right decline by 4.3 points ($p=0.009$, Table~\ref{tab:reversal}). Under B, precision on the 113 paired charts falls from left to right by 5.1 points more than the annotation-sparsity baseline ($p=0.001$), while under R the excess is small and not significant ($-0.8$ points, $p=0.59$). On average, each side loses precision when it moves later in the output. The left side falls from 60.9\% to 58.2\% when written last, and the right side rises from 48.5\% to 51.7\% when written first. A point-level regression on both orders, where output index and $x$ are nearly uncorrelated, estimates precision 2.6 points lower per 100 points written ($p=0.002$, Appendix~\ref{app:controls}).

\paragraph{Robustness and scope.}
The excess is also not significant on the 78 charts where at least 80\% of point pairs were reversed and on the 37 single-series charts that were also at least 80\% reversed, where the reversal is cleanest (Table~\ref{tab:reversal}), and collapse rates are similar (12 of 115 charts under B and 11 under R). Because changing the ordering instruction can also change which points a model writes and how many, this intervention provides evidence that emission order contributes to the decline for GPT-5.6 Luna rather than isolating position alone. The other models did not follow the instruction reliably, so we test them by mirroring the charts instead.

\paragraph{Mirroring the chart instead of the prompt.}
Every model writes points from left to right. We therefore reflect the data in each synthetic scatter plot horizontally, keeping axes, tick labels, and legends in their usual orientation, and give models the same prompt as before. Content on the right of the original chart now appears on the left and is written first, while content on the left is written last. Scoring the first and last quarters of the original ground truth, sorted by $x$, in both versions separates an order effect, the change in precision when the same content is written late rather than early, from a content effect, the difference between right and left content regardless of order. Every model kept writing points in ascending $x$ on mirrored charts.

\begin{table}[H]
\centering
\small
\setlength{\tabcolsep}{3pt}
\resizebox{\columnwidth}{!}{%
\begin{tabular}{lcccccc}
\toprule
& \multicolumn{3}{c}{\textbf{Main}} & \multicolumn{3}{c}{\textbf{Single series}} \\
\cmidrule(lr){2-4}\cmidrule(l){5-7}
\textbf{Model} & $n$ & Order & Content & $n$ & Order & Content \\
\midrule
GPT-5.6 Luna & 197 & $-3.2^{**}$ & $-2.8^{**}$ & 100 & $-5.8^{**}$ & $-3.0^{*}$ \\
Claude Haiku 4.5 & 193 & $-1.3$ & $-4.2^{**}$ & 91 & $-5.8^{**}$ & $-7.4^{**}$ \\
Gemma 4 31B & 192 & $-8.5^{**}$ & $-3.6^{*}$ & 95 & $-10.3^{**}$ & $-5.9^{**}$ \\
Qwen3.8 Flash & 149 & $-18.0^{**}$ & $-1.4$ & 54 & $-23.2^{**}$ & $+9.2$ \\
Fine-tuned & 149 & $-9.1^{**}$ & $-0.6$ & 71 & $-23.9^{**}$ & $+2.2$ \\
\bottomrule
\end{tabular}}
\caption{Order effect (late minus early) and content effect (right minus left) on precision, in percentage points, from mirrored synthetic scatter plots. $n$ counts charts with complete outputs and predictions in every region. $^{*}p<0.05$, $^{**}p<0.01$, Wilcoxon signed-rank.}
\label{tab:mirror}
\end{table}

\paragraph{The same content is less precise when written late.}
On the 200 benchmark scatter plots, the order effect is negative for all five models and significant for four, from $-3.2$ points for GPT-5.6 Luna to $-18.0$ points for Qwen3.8 Flash (Table~\ref{tab:mirror}). The swap is diluted here because each series in a multi-series chart restarts from the left, so content moves only from about 40\% to 60\% of the way through the output. On the 100 additional single-series charts, content moves from about 12\% to 85\% of the way through, and writing it late lowers precision by 5.8 to 23.9 points in every model (all $p\leq0.002$). Right-side content is also less precise in three zero-shot models, by 3.0 to 7.4 points, but the order effect is larger in every model except Claude Haiku 4.5, and the fine-tuned model shows no significant content effect. In point-level regressions on both versions, where output index and $x$ are only weakly correlated, precision falls with output index in all five models, by 1.0 to 2.8 points per 100 points written (Appendix~\ref{app:mirror}).

\section{Long Outputs Collapse and Fabricate}
\label{sec:collapse}

\begin{figure}[t]
\centering
\includegraphics[width=\columnwidth]{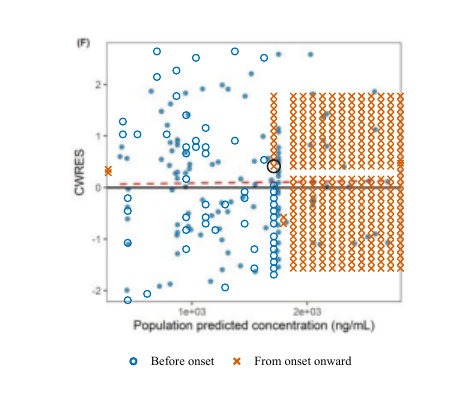}
\caption{A collapsed output from the fine-tuned model on a real scatter plot. Points written before the collapse flag (blue circles) follow the data, while points from the flagged onset onward (orange crosses) form an evenly spaced grid that fills the right of the plot regardless of where the markers lie.}
\label{fig:collapse}
\end{figure}

Alongside the gradual decline, outputs can fail abruptly. Partway through, a model stops tracking the image and writes points that follow a pattern of its own, which we call collapse (Figure~\ref{fig:collapse}). It appears as repeated blocks, series copied from another series, regular grids, and arithmetic progressions, and it can occur in well-formed YAML with plausible values, which schema validation does not catch.

\begin{table}[t]
\centering
\small
\setlength{\tabcolsep}{4pt}
\begin{tabular}{lcccc}
\toprule
& \multicolumn{2}{c}{\textbf{Real}} & \multicolumn{2}{c}{\textbf{Synthetic}} \\
\cmidrule(lr){2-3}\cmidrule(l){4-5}
\textbf{Model} & \textbf{B} & \textbf{E} & \textbf{B} & \textbf{E} \\
\midrule
GPT-5.6 Luna & 10.4 & 19.1 & 1.0 & 7.5 \\
Claude Haiku 4.5 & 42.6 & 58.3 & 19.5 & 31.0 \\
Gemma 4 31B & 28.7 & 39.1 & 8.5 & 12.0 \\
Qwen3.8 Flash & 71.4 & 77.3 & 48.4 & 48.6 \\
Fine-tuned & \multicolumn{2}{c}{62.6} & \multicolumn{2}{c}{30.0} \\
\bottomrule
\end{tabular}
\caption{Share of scatter-plot outputs flagged as collapsed (\%), under the baseline prompt (B) and the prompt stating a lower bound on the number of points (E). The fine-tuned model uses its own prompt.}
\label{tab:collapse}
\end{table}

\paragraph{Collapse is common and starts early.}
Under the baseline prompt, collapse affects 10.4\% of real scatter outputs for GPT-5.6 Luna and 71.4\% for Qwen3.8 Flash (Table~\ref{tab:collapse}). The fine-tuned model collapses on 62.6\%, mostly through grids and progressions, and 22 of its 27 outputs that reach the token limit are collapsed. Its median collapsed output departs from the image within the first 8\% of its points.

\paragraph{Longer outputs collapse more often.}
The extended prompt E lengthens outputs, for example from a median of 126 to 163 points for GPT-5.6 Luna on real scatter plots, and collapse rises with it for every zero-shot model on real charts. Among charts that collapsed under only one prompt, most collapsed under E for GPT-5.6 Luna (15 of 20, $p=0.04$), Claude Haiku 4.5 (27 of 36, $p=0.004$), and Gemma~4~31B (20 of 28, $p=0.04$, exact binomial tests, unadjusted). Only the Claude Haiku 4.5 result remains significant after Holm correction. On synthetic charts, collapse also rises clearly under E for every model except Qwen3.8 Flash, whose rate stays near 48\%.

\paragraph{Models write points where there is no data.}
In 60 synthetic scatter plots, a rectangular region contains no points, and every model writes points into it. Claude Haiku 4.5 does so on 48 of 60 charts, Qwen3.8 Flash writes 2{,}066 such points, and the fine-tuned model 779. GPT-5.6 Luna, the most precise model, writes only 9 points into empty regions under B but 651 under E. Only a handful of these points lie within the matching tolerance of a marker just outside the region. A single sentence asking for more points is enough to make a strong model write points into empty regions that are indistinguishable in format from real extractions.

\section{Chunking Improves Completeness but Not Faithfulness}
\label{sec:chunking}

If precision falls because outputs grow long, extracting a chart in shorter pieces should help. We test this with the fine-tuned model. Each chart is split into four vertical strips or a grid of four by three tiles, and the model is run once per region on an image in which the rest of the plot area is painted white, while axes, labels, and legends remain visible. Plot areas are located from axis lines and reviewed by hand, and region outputs are merged per chart (Appendix~\ref{app:chunking}).

\begin{table}[t]
\centering
\small
\setlength{\tabcolsep}{3.5pt}
\begin{tabular}{lccccc}
\toprule
& & \textbf{Tokens} & & & \textbf{Within call} \\
\textbf{Extraction} & \textbf{Calls} & \textbf{per call} & \textbf{P} & \textbf{R} & \textbf{0--49 / 100--199} \\
\midrule
Whole chart & 1 & 2{,}271 & 46.5 & 48.0 & 51.3 / 34.5 \\
Strips & 4 & 953 & 48.4 & 62.3 & 51.4 / 32.6 \\
Tiles & 12 & 517 & 44.5 & 66.6 & 42.5 / 24.0 \\
\bottomrule
\end{tabular}
\caption{The fine-tuned model extracting whole charts or masked regions of real scatter plots. Tokens per call is the median output length of scatter-plot calls. Precision (P) and recall (R) are computed on the 42 charts complete under all three extractions. Within call gives precision on points 0 to 49 and 100 to 199 of each call's own output, on charts whose calls reach both ranges. On average, precision falls within calls at every granularity ($p<10^{-5}$).}
\label{tab:chunking}
\end{table}

\paragraph{Shorter calls recover more of the data.}
Chunking cuts the median scatter-plot output from 2{,}271 tokens per call to 953 for strips and 517 for tiles, and recall rises from 48.0\% to 62.3\% and 66.6\% ($p<10^{-4}$, Table~\ref{tab:chunking}). Overall precision does not change significantly.

\paragraph{Precision decays again within chunks.}
Within each call's own output, precision still falls on average between the first 50 points and points 100 to 199, from 51.3\% to 34.5\% for whole charts, 51.4\% to 32.6\% within strips, and 42.5\% to 24.0\% within tiles ($p<10^{-5}$, paired within charts). A tile covers one twelfth of the plot area, yet its output follows a similar trajectory to a whole chart. The decline reappears within calls at every granularity, so a point's accuracy depends largely on how far into the output it appears rather than on how much of the chart the call covers.

\paragraph{Collapse persists, and the model writes into masked regions.}
Chunking does not substantially reduce collapse (23 whole-chart outputs collapsed on 42 common charts, against 28 tiled outputs, or 21 after removing points outside each tile). Of 78 tiles whose interior is entirely white, the model wrote points in 49, for 850 points in total, although some may reflect markers just outside a tile.

\paragraph{No detected effect of history length.}
When the fine-tuned model continued pre-written correct answers with 20 to 200 points of history, precision on the next 50 points showed no significant change with history length ($p=0.71$, Appendix~\ref{app:prefix}). Together with the decline inside chunks, this suggests that precision falls with the length of the model's own run of output rather than with the total length of the answer.

\section{Attention to the Image Weakens over Long Outputs}
\label{sec:attention}

We record how the fine-tuned model divides attention between image tokens, prompt text, and previously generated tokens for each generated token, averaged over heads in its ten full-attention layers, on 50 real and 100 synthetic charts.

\paragraph{Attention to the image declines as the output grows.}
On real charts, attention to the image falls over the output in all 50 charts, while attention to generated tokens rises in all 50. Part of this reflects growing generated text competing for attention, but image attention also falls as a share of attention to the prompt in 38 of 50 charts.

\begin{figure}[H]
\centering
\includegraphics[width=\columnwidth]{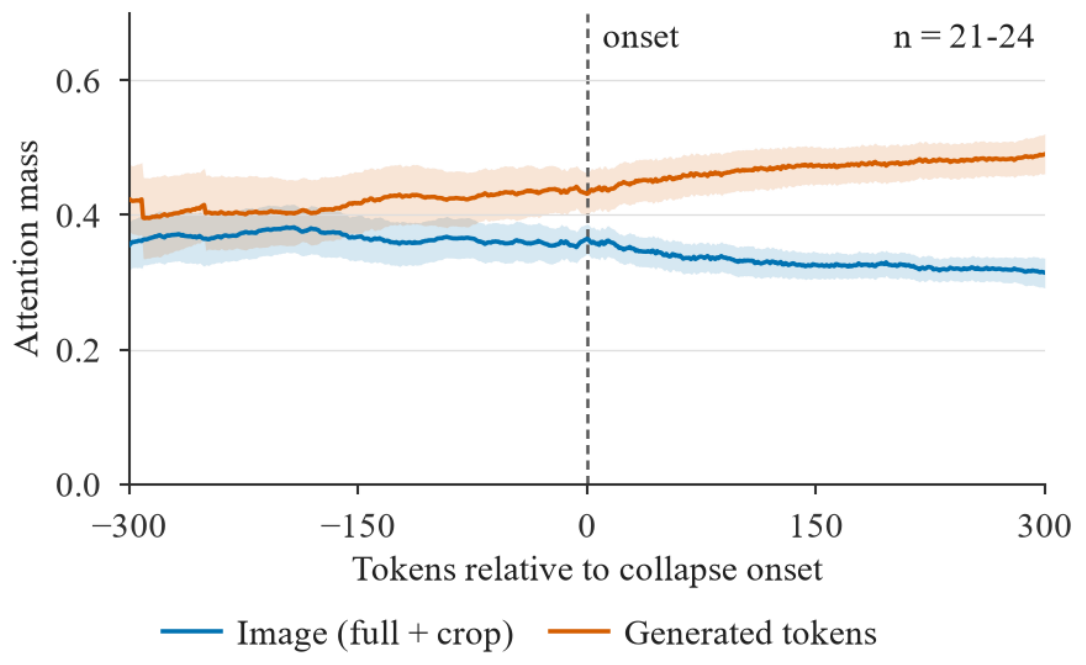}
\caption{Attention in the fine-tuned model around collapse onset, averaged over 21 to 24 collapsed synthetic scatter plots aligned at the first collapsed point. After onset, attention to the image is lower and attention to previously generated tokens higher, although the trajectories also reflect the general decline along the output.}
\label{fig:attention}
\end{figure}

\paragraph{Collapse-flagged segments attend less to the image.}
On synthetic scatter plots, within the 50-point bin of the output that contains collapse onset, collapse-flagged points receive 5.8 percentage points less image attention and 7.3 percentage points more attention to generated tokens than the preceding points (22 of 24 charts, $p<10^{-5}$, Figure~\ref{fig:attention}). The difference grows with bin width, and although it partly reflects the general decline along the output, collapse-flagged segments rely less on the image and more on the model's own output.

\paragraph{No robust link to individual errors.}
At a fixed output position, image attention is not robustly associated with whether an individual point is correct ($-0.1$ percentage points with fine position matching, $p=0.50$). Attention thus tracks the course of an output but does not identify individual errors.

\section{Discussion}
\label{sec:discussion}

\paragraph{Output-side grounding.}
Long-context evaluation usually asks whether a model can use a long input. Our results concern the reverse case, keeping a long output faithful to one fixed image, and every model is more precise on the first points of a dense chart than on the last. Reading quality is therefore not a fixed property of a model and an image, and chart benchmarks built on short answers or chart-level scores do not measure it.

\paragraph{Two failures, one direction.}
Precision erodes gradually, and outputs can collapse abruptly. Both grow more likely as outputs lengthen, and in the fine-tuned model attention to the image falls over the same span, most clearly in collapse-flagged segments. This fits accounts in which later tokens depend increasingly on the model's own prefix \citep{chung_v1_2025, favero_multimodal_2024} and errors compound from it \citep{bengio_scheduled_2015}. Neither fine-tuning, prompting for completeness, nor chunking removes the failure, which suggests changes to training or decoding.

\paragraph{Implications for extraction pipelines.}
Collapsed and fabricated outputs are well-formed, so extraction pipelines would store them as real data. Prompts demanding a minimum number of points should be avoided, and shorter calls improve recall. Post-hoc filters, such as keeping points that a second run confirms, screening with the collapse detector, or asking the model to check its own points, raise precision only partly and lose recall (Appendix~\ref{app:mitigation}).

\section{Conclusion}
\label{sec:conclusion}

We used chart digitization to test whether VLMs stay faithful to an image over long outputs. Across five models, precision on dense scatter plots declines with output position beyond what incomplete annotation explains, and two ordering interventions provide evidence that emission order contributes. Long outputs also collapse and fabricate points in empty regions, and because these errors are well-formed, faithfulness over long outputs must be measured directly.

\section*{Limitations}

\paragraph{Causal evidence.}
The reversed-order test requires a model to follow an instruction to write points from right to left, and only GPT-5.6 Luna complied reliably. Mirroring synthetic charts extends the test to all five models without changing the prompt, but it moves content to the other side of the image as well as to the other end of the output, and fixed legends hide slightly different markers. Both tests therefore provide evidence that emission order contributes rather than isolating it, and the mirroring test uses synthetic charts only. Qwen3.8 Flash and the fine-tuned model contribute fewer mirrored charts because more of their outputs are truncated or malformed.

\paragraph{Annotation coverage.}
Real annotations record a representative subset of points in dense clouds. We address this in two ways, with a split-half baseline that estimates the effect of annotation density, and with a synthetic benchmark whose ground truth is exhaustive. Both confirm the decline, although synthetic charts are cleaner than real figures, so we compare the two settings descriptively.

\paragraph{Fine-tuned model.}
The fine-tuned model uses its own prompt and inference settings, so comparisons with the zero-shot models reflect each model together with its setup. The chunking experiments and the attention analysis were run on the fine-tuned model, because it can be run locally with direct access to attention weights.

\paragraph{Attention analysis.}
Attention describes where a model looks rather than what determines its predictions \citep{jain_attention_2019, serrano_attention_2019}, so we interpret the attention results as consistent with, rather than proof of, a shift in reliance from the image to generated text.

\paragraph{Scope.}
Our real charts come from biomedical articles, and we focus on scatter plots and line charts, where outputs are longest. The same analysis applies directly to other chart types and domains. The annotated real charts used in our evaluation are available upon request for research use, and we will release our code, the collapse detector, and the synthetic benchmark generator upon publication.

\bibliography{references}

\appendix
\raggedbottom

\section{Dataset and Preprocessing}
\label{app:dataset}
\label{app:pipeline}
The fine-tuned model was trained on a human-annotated dataset of 3{,}628 multi-panel figures from open-access PubMed Central (PMC) articles. The figures were split manually into 22{,}285 individual charts, which served as its training data.

\begin{figure}[h]
    \centering
    \includegraphics[width=\columnwidth]{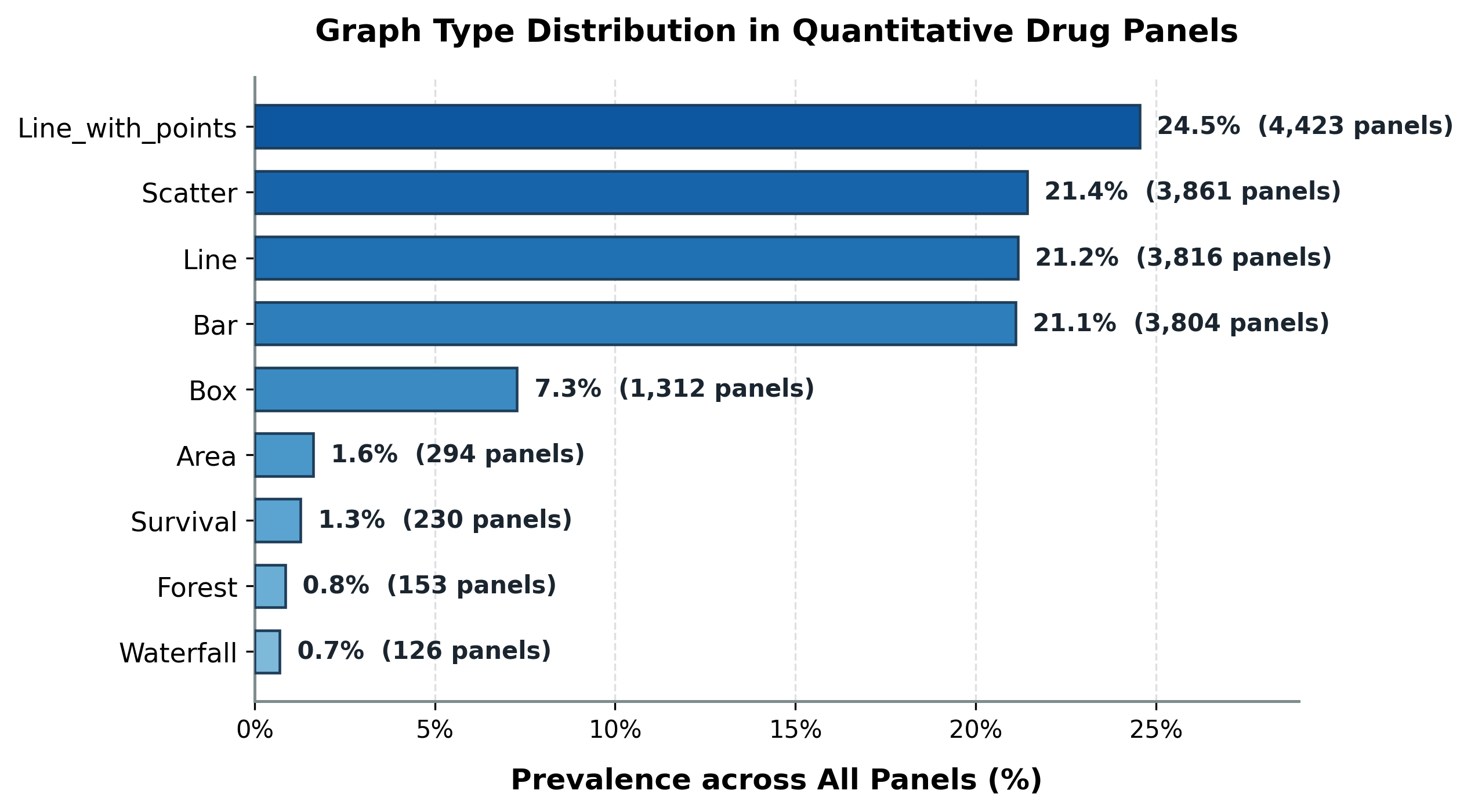}
    \caption{Distribution of the most common graph types in the training set.}
    \label{fig:graph_distribution}
\end{figure}

We focused our annotation efforts on the most prevalent quantitative graph types found in scientific literature. As illustrated in Figure~\ref{fig:graph_distribution}, the dataset predominantly consists of line graphs (with and without points), scatter plots, bar charts, and box plots, along with a long tail of specialized biomedical charts (e.g., area, survival, forest, and waterfall plots).

\subsection{Annotation Methodology and Availability}

Annotation was managed through an internal web-based platform. Figures were first pre-annotated by a general-purpose vision-language model, and a team of seven annotators then reviewed and corrected the structural metadata and the coordinate points. Inter-annotator agreement was monitored, and disagreements were resolved by consensus review. Pre-annotation could leave some regions of a chart less completely annotated than others, which is one reason we use the annotation-sparsity null and the synthetic benchmark with exhaustive ground truth.

A further 400 figures were held out for the final evaluation dataset. To prevent data leakage, we enforced a document-level split, ensuring that no figures from the same PMC article were present in both the training and test sets. The full training dataset and its annotations are proprietary and are not publicly released. The 175 evaluation panels and their annotations are available upon request for research use.

\section{YAML Extraction Schema}
\label{app:schema}

The ground-truth targets use a nested YAML schema that separates panel-level metadata, such as axis labels, ranges, ticks, and scales, from series-level data. Each series is linked to a legend entry and lists its data points as coordinate pairs, with optional upper and lower error bars given as coordinate pairs in the same format. Box plots add fields for quartiles, whiskers, and outliers.

\begin{lstlisting}
panels:
  - panel_id: "single"
    plot_type: "line"
    panel_title: ""
    legend:
      - id: "series_1"
        label: "Series 1"
        color: "black line with circle markers"
    x_axis:
      label: ""
      scale: "linear"
      range: [0, 1]
      ticks: [0, 0.5, 1]
    y_axis:
      label: ""
      scale: "linear"
      range: [0, 1]
      ticks: [0, 0.5, 1]
    series:
      - legend_id: "series_1"
        data_points:
          - [0.0, 0.1]
          - [1.0, 0.9]
          - ...
        error_bars:
          upper:
            - [0.0, 0.15]
            - ...
          lower:
            - [0.0, 0.05]
            - ...
        observation: "Rises overall with a mid-range dip."
    panel_observation: "Black line with circle markers corresponds to Series 1."
\end{lstlisting}

\section{Fine-Tuning Details}
\label{app:hyper}

The fine-tuned model was trained with QLoRA. Trainable low-rank adapters were inserted into the vision and language modules, including attention and MLP layers, while the base weights were kept frozen in 4-bit NormalFloat (NF4) precision, which allowed training on a single 80GB NVIDIA H100 GPU. Table~\ref{tab:hyperparams} lists the settings.

\begin{table}[h]
    \centering
    \small
    
    \begin{tabular}{ll}
        \toprule
        \textbf{Parameter} & \textbf{Value} \\
        \midrule
        Hardware & 1$\times$ NVIDIA H100 \\
        Training time & $\sim$36 hours \\
        Epochs & 1 \\
        Effective batch size & 16 \\
        Learning rate & $2\times 10^{-5}$ \\
        LR scheduler & Cosine \\
        LoRA rank ($r$) & 32 \\
        LoRA alpha & 32 \\
        Target modules & Vision, language, attention, MLP \\
        Quantization & 4-bit (Unsloth) \\
        \bottomrule
    \end{tabular}
    \caption{Fine-tuning hyperparameters.}
    \label{tab:hyperparams}
\end{table}

\section{Synthetic Benchmark}
\label{app:synthetic}

The synthetic benchmark contains 300 charts, 200 scatter plots and 100 line charts, generated with a fixed random seed of 42 and rendered with Matplotlib in styles modeled on biomedical figures. Ground truth uses the same YAML schema as the real annotations, allowing both datasets to share the existing extraction pipeline and core scoring functions.

\paragraph{Scatter plots.}
The scatter plots are split into three density tiers by total point count, 70 charts with 50 to 100 points, 70 with 100 to 250, and 60 with 250 to 600. Each chart uses one of five layouts, 40 charts each. Random clouds and clusters place points freely or around a few centers. Observed-versus-predicted plots scatter points around the identity line, and residual plots scatter them around zero. Pharmacokinetic stacks place many points at a few shared sampling times, which produces dense vertical columns that are common in pharmacology figures and hard to read.

\paragraph{Line charts.}
The line charts contain two to six series with 10 to 60 points each and follow four layouts, 25 charts each, namely dose-response curves, growth and decay curves, pharmacokinetic concentration curves, and repeated-dose profiles.

\paragraph{Axes and styles.}
Ninety charts use a logarithmic $y$-axis and 30 a logarithmic $x$-axis. Styles are drawn at random per chart, including font family and size, color palette or grayscale, background, gridlines, marker shapes and sizes, hollow or filled markers, line styles, legend placement and framing, confidence bands, error bars, and text annotations. Half of the charts are embedded as one panel of a multi-panel figure, and models receive the full figure together with the target panel, as for real charts.

\paragraph{Ground truth.}
Every plotted point is recorded in data coordinates together with its pixel position, for 57{,}793 points in total (median 154.5 per chart). A marker counts as fully hidden when no pixel of its rendered silhouette remains visible after later markers and opaque legends are drawn, using an alpha threshold of 16 out of 255. This flags 3{,}420 markers, which remain in the primary metrics and are removed only for visible-only recall and for the precision robustness check. In 60 scatter plots, a rectangle in data coordinates contains no ground-truth marker centers. Its bounds are stored with the chart, and predictions inside it are recorded as empty-region violations. Predictions near the boundary may nevertheless fall within the matching tolerance of a marker outside the region. We checked rendered overlays of ground truth on example charts and validated the full manifest automatically, with no inconsistencies.

\section{Collapse Detector}
\label{app:detector}

The detector checks each output for four patterns and records the index at which each begins. Coordinates are compared exactly or after rounding to six decimal places, as noted below, and a single qualifying pattern is enough to flag an output.

\paragraph{Repeated blocks.}
A block of 15 consecutive points, taken in emission order across the whole output, reproduces an earlier block of 15 points, after rounding both coordinates to six decimal places. The block must contain at least five distinct coordinate pairs. Matching preserves point order and can cross series boundaries, but does not detect copies shifted by a constant offset.

\paragraph{Duplicated series.}
A series with at least five points is flagged when at least 80\% of its points also appear in an earlier series in the same output, after rounding to six decimal places. Overlap counts repeated points with their multiplicity and ignores point order, and shifted copies are not detected.

\paragraph{Regular grids.}
The grid check detects evenly spaced columns, runs of points at a single $x$ value whose $y$ values advance in equal steps. In linear space, it requires eight consecutive points in one series whose $x$ values agree within $10^{-9}$ and whose $y$ steps are equal within $10^{-9}$ and nonzero. A second check applies the same test in logarithmic space to five consecutive points with positive $y$ values, requiring equal steps in $\log_{10} y$. This check runs regardless of the declared axis scale.

\paragraph{Arithmetic progressions.}
A window of 20 consecutive points in one series is flagged when both its $x$ steps and its $y$ steps repeat with a period of one to eight points, possibly different for the two coordinates, and the window contains at least five distinct coordinate pairs. Steps are compared with a tolerance of $10^{-6}$ times the largest absolute coordinate in the series, or $10^{-6}$ if that is smaller than one, and windows in which $y$ does not change are excluded. Because many line charts are sampled at evenly spaced $x$ values, progressions alone do not count as collapse on line charts, although the other three patterns can still flag them.

\paragraph{Onset and denominators.}
The onset of a flagged output is the earliest point index at which a counted pattern begins, namely the start of the later repeated block, of the duplicated series, or of the first qualifying grid or progression window. It marks where the pattern starts rather than where enough evidence has accumulated to detect it. The fractional onset divides this index by the total number of recovered points. Collapse rates include every output with at least one recoverable point, including outputs that reached the token limit and outputs recovered from partly malformed responses. Missing outputs and outputs with no recoverable points are excluded.

\paragraph{Validation and interpretation.}
Applied to the exact ground-truth point sets of all 300 synthetic charts, including pharmacokinetic stacks and evenly sampled line charts, the detector flags none. A flag marks structural degeneration rather than a confirmed error. Approximate coordinates can occasionally look regular while remaining correct, and in one synthetic chart all 109 predicted points fall within the matching tolerance yet the output is flagged as a grid.

\section{Chunked Extraction}
\label{app:chunking}

\paragraph{Locating the plot area.}
For each panel, we locate the plot area from axis lines in the image, taking rows and columns in which at least 40\% of pixels are dark (grayscale value below 180) as candidate axes. When no axes are found, a default box covering 15\% to 96\% of the width and 4\% to 85\% of the height is used. Every plot area was reviewed on contact sheets and corrected by hand where needed. Ground-truth annotations are never used, so the same procedure is available to a model at inference time.

\paragraph{Masking.}
The plot area is divided into four vertical strips of equal width, or into a grid of four columns and three rows. For each region, everything inside the plot area outside that region is painted white, while axes, tick labels, axis titles, legends, and panel labels outside the plot area are left unchanged. The same mask is applied to the target panel within the full figure, located by template matching, so the hidden data cannot be read from the second image. Masks were verified for all 700 strip images and 2{,}100 tile images. Markers that straddle a region boundary may remain partly visible in adjacent regions.

\paragraph{Inference and merging.}
Each region is a separate call with the fine-tuned model's usual prompt and settings, including its 8{,}000-token limit. Of 700 strip calls and 2{,}100 tile calls, 30 and 73 reached the limit. Region outputs are merged per chart by grouping series by legend label, after normalizing case and whitespace, and concatenating their points left to right for strips and row by row for tiles. Exact coordinate duplicates between horizontally or vertically adjacent regions within the same series are removed, while duplicates within a region or between diagonal and nonadjacent regions are kept. Merged outputs are then scored exactly like whole-chart outputs. Our primary results use these unfiltered merges. As a sensitivity check, a filtered variant also removes points that the model placed outside its region, judged using the model's own reported axes rather than ground truth. It takes the median lower and upper axis bounds over valid region outputs that share the majority axis scale, maps each region's pixel bounds into data coordinates with a margin of 5\% of the full axis range on each side, and filters strips by $x$ only and tiles by both $x$ and $y$. On logarithmic axes, mapping and margins are computed in $\log_{10}$ space and nonpositive coordinates are rejected. The filtered variant leads to the same conclusions. The overall precision and recall in Table~\ref{tab:chunking} use the 42 charts for which all three extractions produced a complete output. The within-call comparison uses a different set of charts, those in which a call wrote more than 100 points. It compares whatever points the call wrote in positions 100 to 199 with points 0 to 49 of the same call. This gives 53 charts for whole-chart extraction, 55 for strips, and 62 for tiles.

\paragraph{Blank regions.}
A tile counts as blank when every pixel inside its region is white, which identifies 78 tiles. Because no data is visible inside these regions, points written there cannot come from reading markers within the tile, independent of how pixels are mapped to data coordinates, although some may reflect markers just outside its boundary.

\section{Forced-Prefix Test}
\label{app:prefix}

We generated 60 additional synthetic scatter charts with the same generator, each with a single series of 300 points, spread evenly over the random cloud, cluster, observed-versus-predicted, and residual layouts. For each chart, we wrote the beginning of the fine-tuned model's answer in its own output format, including the header and the last $h$ of the first 200 ground-truth points in ascending order of $x$, rounded as the model rounds its own coordinates, and let the model continue. We used $h = 20$, 50, 150, and 200, giving 240 continuations with the same continuation point per chart. Each continuation was limited to 1{,}500 output tokens. Of the 240 continuations, 238 reached this limit. All 240 were retained, since every continuation contained at least 50 complete points. We scored the first 50 generated points at $d=0.02$ against the 100 remaining ground-truth points.

All 240 continuations started at or to the right of the last pre-written point. Mean precision was 56.7\%, 56.8\%, 58.9\%, and 58.0\% for $h = 20$, 50, 150, and 200. The slope of precision against history length, with chart fixed effects, was $+1.0$ points per 100 points of history (95\% confidence interval $-4.4$ to $+6.4$, $p=0.71$), and no pairwise difference was significant. The collapse detector flagged 27, 35, 33, and 31 of 60 continuations within their first 50 points, with a median onset between the sixth and ninth generated point. Excluding these continuations left the trend non-significant ($p=0.25$).

\section{Mirrored Charts}
\label{app:mirror}

\paragraph{Construction.}
For each of the 200 synthetic scatter plots, we regenerate the chart with the same generator settings and reflect the target panel's data horizontally, $x' = x_{\min} + x_{\max} - x$, computed in $\log_{10}$ space on logarithmic axes. Axes, tick labels, legends, and other panels keep their original orientation and position, and the empty-region rectangle is reflected with the data. Pixel positions and hidden markers are recomputed from the new render with the same alpha threshold. Models receive the mirrored full figure and panel crop with the same prompts and settings as before, and outputs for the original charts are reused. Predictions on mirrored charts are reflected back before scoring, which leaves the matching unchanged.

\paragraph{Additional single-series charts.}
Only 21 benchmark scatter plots have a single series. We therefore generate 100 more with a separate seed, 34 with 50 to 100 points and 33 each with 100 to 250 and 250 to 600 points, 20 per layout, with no empty region and the same style randomization, and render each in original and mirrored form. None duplicates a benchmark chart. For these charts, Gemma~4~31B is routed to a single provider in both versions. Pooling them with the benchmark single-series charts gives order effects from $-7.0$ to $-24.8$ points across the five models.

\paragraph{Effects.}
Content regions are the first and last quarters of the original ground-truth points sorted by $x$, with ties kept. A chart enters the analysis when both outputs are complete and every region contains at least one prediction in both versions. Effects are means over charts with 95\% bootstrap confidence intervals from 2{,}000 resamples.

\paragraph{Regression.}
For each model, we fit a logistic regression on the pooled original and mirrored outputs of the 200 benchmark scatter plots, with one row per predicted point and correctness at $d=0.02$ as the outcome. The predictors are output index, the point's $x$ in the original and in the displayed chart, its normalized $y$, local marker density, and an indicator for the mirrored version, with a shared intercept for each chart and its mirror and standard errors clustered by chart. Variance inflation factors for output index are at most 1.25.

\begin{table}[H]
\centering
\small
\setlength{\tabcolsep}{4pt}
\begin{tabular}{lcc}
\toprule
\textbf{Model} & \textbf{Odds ratio (95\% CI)} & \textbf{AME} \\
\midrule
GPT-5.6 Luna & 0.83 (0.77 to 0.88) & $-2.3$ \\
Claude Haiku 4.5 & 0.93 (0.91 to 0.96) & $-1.0$ \\
Gemma 4 31B & 0.91 (0.88 to 0.93) & $-1.6$ \\
Qwen3.8 Flash & 0.88 (0.84 to 0.92) & $-1.7$ \\
Fine-tuned & 0.82 (0.74 to 0.90) & $-2.8$ \\
\bottomrule
\end{tabular}
\caption{Effect of output position per 100 points on the probability that a point is correct, from regressions on original and mirrored synthetic scatter plots. AME is the average marginal effect in percentage points. All odds ratios are significant ($p<10^{-4}$).}
\label{tab:app_mirror}
\end{table}

\section{Post-hoc Mitigations}
\label{app:mitigation}

We tested three filters on existing outputs, without new model calls (Table~\ref{tab:app_mitigation}). The agreement filter keeps a point only if a second extraction of the same chart contains a point within $d=0.02$. For GPT-5.6 Luna, the second extraction is the reversed-order run. For the fine-tuned model, it is the strip or tile extraction. The order merge takes points left of the median predicted $x$ from the left-to-right run and the remaining points from the right-to-left run. Detector screening removes all points from the first collapse onset onward. We also tested self-verification on the 200 synthetic scatter plots, using new calls in which each model received its own extracted points in batches of 50, with the same images, and marked each point as kept or removed. Precision rose by 2.0 points for GPT-5.6 Luna and 3.0 points for Qwen3.8 Flash, but Claude Haiku 4.5 and Gemma~4~31B removed correct and incorrect points at nearly the same rate, lowering recall by 15 and 19 points without improving precision. Verification removed every point from 30 Claude Haiku 4.5 charts and 57 Gemma~4~31B charts, which are left out of the precision comparison and count as zero recall. The decline with output position remained significant after verification for all four models.

Agreement raises precision by 5.5 to 10.3 points and lowers recall by 3.5 to 14.1 points, on the 113 paired GPT-5.6 Luna charts and the 42 charts with all three extractions from the fine-tuned model. Its precision exceeds that of the same filter applied with extractions of unrelated charts by 4.7 to 8.9 points ($p<0.001$), averaged over 200 random reassignments of charts with each extraction normalized by its own chart's axes, so the gain does not arise from chance proximity in dense charts. The order merge does not change precision significantly. Screening helps most where collapse is most common, but it can remove an entire output, as for 11 of the 85 complete real outputs of the fine-tuned model. Precision is then undefined, and those charts are left out of the precision comparison. On the synthetic charts with an empty region, counted over all recoverable outputs as in Section~\ref{sec:collapse}, it reduces the points written there from 2{,}066 to 142 for Qwen3.8 Flash, from 779 to 283 for the fine-tuned model, from 639 to 346 for Claude Haiku 4.5, and from 651 to 474 for GPT-5.6 Luna under the extended prompt. All three filters remove points rather than correct them. A decline with output position remains after agreement filtering, for GPT-5.6 Luna ($-11.8$ points from the first to the last original quarter, $p<0.001$) and for the fine-tuned model ($-8.6$ points, $p=0.04$).

\begin{table}[H]
\centering
\small
\setlength{\tabcolsep}{3.5pt}
\begin{tabular}{llcc}
\toprule
\textbf{Filter} & \textbf{Model} & $\Delta$\textbf{P} & $\Delta$\textbf{R} \\
\midrule
Agreement, reversed run & GPT-5.6 Luna & $+5.8$ & $-14.1$ \\
Order merge & GPT-5.6 Luna & $+1.0^{\dagger}$ & $-1.5^{\dagger}$ \\
Agreement, either & Fine-tuned & $+5.5$ & $-3.5$ \\
Agreement, both & Fine-tuned & $+10.3$ & $-12.3$ \\
\midrule
Detector screening & GPT-5.6 Luna & $+0.6^{\dagger}$ & $-5.4$ \\
                   & Claude Haiku 4.5 & $+1.7^{\dagger}$ & $-5.1$ \\
                   & Gemma 4 31B & $+3.0$ & $-6.7$ \\
                   & Qwen3.8 Flash & $+9.4$ & $-12.3$ \\
                   & Fine-tuned & $+3.6$ & $-18.8$ \\
\bottomrule
\end{tabular}
\caption{Change in precision and recall (percentage points) on real scatter plots after post-hoc filtering, relative to the unfiltered output. Zero-shot models use the baseline prompt. Either and both refer to agreement with the strip or tile extraction, or with both. Precision changes are paired over charts that retain at least one point. $^{\dagger}$ Not significant ($p>0.05$, Wilcoxon signed-rank).}
\label{tab:app_mitigation}
\end{table}

\section{Additional Controls}
\label{app:controls}

\paragraph{Marker exhaustion.}
A model could lose precision late in its output because it has no markers left to read. On synthetic scatter plots, we therefore restricted the analysis to outputs that contain no more points than the chart. The first-to-last quarter decline remains significant in 8 of 9 model and prompt groups ($-3.9$ to $-8.7$ points). The exception is GPT-5.6 Luna under the extended prompt, where only 22 outputs qualify. Restricting further to outputs with at most 80\% as many points as the chart has visible markers, the decline is significant in all 8 groups that can be tested ($-2.3$ to $-10.6$ points). In outputs that do exceed the chart's point count, points beyond that count are less precise than earlier ones for every model.

\paragraph{Point-level regression.}
To separate output position from where a point lies, we fit a logistic regression per model, prompt, and dataset, with one row per predicted point. The outcome is whether the point is correct at $d=0.02$. The predictors are the point's index in the output, its normalized $x$ and $y$, and the logarithm of one plus the number of ground-truth points within 0.05 of it, with a separate intercept for each chart and standard errors clustered by chart. Output index has a significant negative coefficient in 13 of 18 fits (Table~\ref{tab:app_controls}). The exceptions are all four fits for Claude Haiku 4.5 and Gemma~4~31B under the extended prompt on real charts. With position expressed as a fraction of the output length, the coefficient is significant in 8 of 18 fits. For GPT-5.6 Luna on real charts, a joint fit on both emission orders, in which output index and $x$ are nearly uncorrelated within charts ($r=0.08$), gives an odds ratio of 0.86 per 100 points (95\% confidence interval 0.78 to 0.94, $p=0.002$), an average marginal effect of $-2.6$ points per 100 points. This fit uses the 110 of 113 paired charts whose points are not all correct or all incorrect. Local density depends on the ground truth and is mechanically related to correctness, so these fits adjust for it without treating it as a cause.

\begin{table}[H]
\centering
\small
\setlength{\tabcolsep}{4pt}
\begin{tabular}{llcc}
\toprule
\textbf{Model} & & \textbf{Real} & \textbf{Synthetic} \\
\midrule
GPT-5.6 Luna & B & $-2.3$ & $-2.2$ \\
             & E & $-2.7$ & $-1.6$ \\
Claude Haiku 4.5 & B & $+0.4^{\dagger}$ & $-0.7^{\dagger}$ \\
                 & E & $-0.3^{\dagger}$ & $+1.0^{\dagger}$ \\
Gemma 4 31B & B & $-3.6$ & $-0.9$ \\
            & E & $-0.3^{\dagger}$ & $-1.2$ \\
Qwen3.8 Flash & B & $-1.1$ & $-1.1$ \\
              & E & $-0.7$ & $-0.7$ \\
Fine-tuned & & $-2.6$ & $-3.8$ \\
\bottomrule
\end{tabular}
\caption{Average marginal effect of output position on the probability that a point is correct (percentage points per 100 points) on scatter plots, from logistic regressions with chart fixed effects that adjust for the point's location and local marker density. $^{\dagger}$ Not significant ($p>0.05$).}
\label{tab:app_controls}
\end{table}

\section{Additional Results}
\label{app:tables}

Tables~\ref{tab:app_real} and~\ref{tab:app_synth} report precision and recall on scatter plots at three matching tolerances, and Table~\ref{tab:app_tolerance} reports the first-to-last quarter change in precision at the stricter and looser tolerances. Table~\ref{tab:app_line} reports results for line charts, and Table~\ref{tab:app_robust} reports robustness checks. All values are means over charts.

\begin{table*}[!tp]
\centering
\small
\begin{tabular}{llcccccc}
\toprule
& & \multicolumn{3}{c}{\textbf{Precision}} & \multicolumn{3}{c}{\textbf{Recall}} \\
\cmidrule(lr){3-5}\cmidrule(l){6-8}
\textbf{Model} & & $d{=}0.01$ & $d{=}0.02$ & $d{=}0.05$ & $d{=}0.01$ & $d{=}0.02$ & $d{=}0.05$ \\
\midrule
GPT-5.6 Luna & B & 27.2 & 55.2 & 88.8 & 25.9 & 52.8 & 84.2 \\
             & E & 24.7 & 50.4 & 87.2 & 29.4 & 55.1 & 86.0 \\
Claude Haiku 4.5 & B & 13.4 & 29.9 & 65.3 & 8.3 & 19.8 & 50.0 \\
                 & E & 13.3 & 28.6 & 61.9 & 12.1 & 24.8 & 53.8 \\
Gemma 4 31B & B & 22.9 & 44.7 & 83.2 & 13.1 & 28.2 & 65.4 \\
            & E & 19.6 & 41.6 & 79.7 & 16.2 & 35.8 & 72.2 \\
Qwen3.8 Flash & B & 18.1 & 35.6 & 64.7 & 17.4 & 32.7 & 63.8 \\
              & E & 17.6 & 34.0 & 63.2 & 21.0 & 37.6 & 65.2 \\
Fine-tuned & & 23.1 & 47.7 & 81.7 & 22.4 & 45.8 & 83.3 \\
\bottomrule
\end{tabular}
\caption{Precision and recall (\%) on real scatter plots at three matching tolerances, computed over complete outputs. Recall on real charts is measured against subsampled annotations.}
\label{tab:app_real}
\end{table*}

\begin{table*}[!tp]
\centering
\small
\begin{tabular}{llccccccc}
\toprule
& & \multicolumn{3}{c}{\textbf{Precision}} & \multicolumn{3}{c}{\textbf{Recall}} & \textbf{Visible recall} \\
\cmidrule(lr){3-5}\cmidrule(lr){6-8}\cmidrule(l){9-9}
\textbf{Model} & & $d{=}0.01$ & $d{=}0.02$ & $d{=}0.05$ & $d{=}0.01$ & $d{=}0.02$ & $d{=}0.05$ & $d{=}0.02$ \\
\midrule
GPT-5.6 Luna & B & 55.6 & 83.7 & 98.7 & 48.1 & 75.2 & 95.9 & 75.4 \\
             & E & 50.2 & 79.7 & 97.5 & 53.7 & 80.6 & 96.2 & 80.8 \\
Claude Haiku 4.5 & B & 15.7 & 36.2 & 75.8 & 8.1 & 21.6 & 56.9 & 21.6 \\
                 & E & 13.7 & 33.1 & 71.3 & 9.6 & 25.4 & 60.5 & 25.3 \\
Gemma 4 31B & B & 28.4 & 63.4 & 95.9 & 13.3 & 35.8 & 78.8 & 35.9 \\
            & E & 29.1 & 63.3 & 95.0 & 16.1 & 39.9 & 81.1 & 40.1 \\
Qwen3.8 Flash & B & 26.3 & 55.0 & 83.6 & 20.1 & 46.3 & 80.3 & 46.3 \\
              & E & 27.2 & 54.8 & 83.6 & 23.3 & 50.3 & 81.0 & 50.3 \\
Fine-tuned & & 41.1 & 73.3 & 95.3 & 32.0 & 63.0 & 92.2 & 63.0 \\
\bottomrule
\end{tabular}
\caption{Precision and recall (\%) on synthetic scatter plots with exhaustive ground truth. Visible recall excludes markers that are fully hidden by other markers or by opaque legends.}
\label{tab:app_synth}
\end{table*}

\begin{table*}[!tp]
\small
\begin{minipage}[t]{0.48\textwidth}
\centering
\setlength{\tabcolsep}{3.5pt}
\begin{tabular}{llcccc}
\toprule
& & \multicolumn{2}{c}{\textbf{Real}} & \multicolumn{2}{c}{\textbf{Synthetic}} \\
\cmidrule(lr){3-4}\cmidrule(l){5-6}
\textbf{Model} & & $0.01$ & $0.05$ & $0.01$ & $0.05$ \\
\midrule
GPT-5.6 Luna & B & $-5.8$ & $-8.0$ & $-4.7$ & $-0.5$ \\
             & E & $-11.2$ & $-10.4$ & $-5.5$ & $-1.9$ \\
Claude Haiku 4.5 & B & $-3.0$ & $-20.5$ & $-1.1$ & $-7.7$ \\
                 & E & $-3.2$ & $-21.7$ & $-2.5$ & $-9.6$ \\
Gemma 4 31B & B & $-5.8$ & $-14.5$ & $-1.2$ & $-5.5$ \\
            & E & $-8.8$ & $-15.3$ & $-1.8$ & $-6.3$ \\
Qwen3.8 Flash & B & $-9.5$ & $-32.1$ & $-3.4$ & $-12.4$ \\
              & E & $-8.7$ & $-27.0$ & $-5.0$ & $-10.2$ \\
Fine-tuned & & $-5.7$ & $-14.0$ & $-5.2$ & $-3.7$ \\
\bottomrule
\end{tabular}
\caption{Change in mean precision (percentage points) from the first to the last quarter of the output on scatter plots at tolerances $d=0.01$ and $d=0.05$. Mean precision is lower in the last quarter than in the first at every tolerance for every model. At $d=0.05$, GPT-5.6 Luna is close to its ceiling on synthetic charts, which leaves little room for decline. The changes are significant ($p<0.05$, Wilcoxon signed-rank) in 31 of 36 cases. The exceptions are Claude Haiku 4.5 under E on real charts at $d=0.01$, Gemma 4 31B under B and E and Claude Haiku 4.5 under B on synthetic charts at $d=0.01$, and GPT-5.6 Luna under B on synthetic charts at $d=0.05$.}
\label{tab:app_tolerance}
\end{minipage}\hfill
\begin{minipage}[t]{0.48\textwidth}
\centering
\setlength{\tabcolsep}{4pt}
\begin{tabular}{lcccc}
\toprule
& \multicolumn{2}{c}{\textbf{Real}} & \multicolumn{2}{c}{\textbf{Synthetic}} \\
\cmidrule(lr){2-3}\cmidrule(l){4-5}
\textbf{Model} & $\Delta$ & $p$ & $\Delta$ & $p$ \\
\midrule
GPT-5.6 Luna & $+3.9$ & 0.18 & $-2.0$ & 0.56 \\
Claude Haiku 4.5 & $-3.2$ & 0.40 & $-0.7$ & 0.92 \\
Gemma 4 31B & $-2.7$ & 0.41 & $-3.2$ & 0.63 \\
Qwen3.8 Flash & $-3.5$ & 0.64 & $-14.6$ & $<0.001$ \\
Fine-tuned & $+1.2$ & 0.58 & $-12.3$ & $<0.001$ \\
\bottomrule
\end{tabular}
\caption{Change in precision (percentage points) from the first to the last quarter of the output on line charts at $d=0.02$, with zero-shot models under the baseline prompt and the fine-tuned model under its own prompt. Real line-chart outputs are short and show no significant decline. On synthetic line charts, which require longer outputs, precision declines for Qwen3.8 Flash and the fine-tuned model.}
\label{tab:app_line}
\end{minipage}
\end{table*}

\begin{table*}[!tp]
\centering
\small
\begin{tabular}{llcccc}
\toprule
& & \multicolumn{2}{c}{\textbf{One-to-one matching}} & \textbf{Hidden removed} & \textbf{One panel per article} \\
\cmidrule(lr){3-4}\cmidrule(lr){5-5}\cmidrule(l){6-6}
\textbf{Model} & & Real & Synthetic & Synthetic & Real \\
\midrule
GPT-5.6 Luna & B & $-6.1$ & $-4.9$ & $-5.0$ & $-10.8$ \\
             & E & $-7.6$ & $-5.3$ & $-6.4$ & $-15.1$ \\
Claude Haiku 4.5 & B & $-7.9$ & $-4.5$ & $-5.0$ & $-9.7$ \\
                 & E & $-4.9$ & $-5.2$ & $-6.0$ & $-10.9$ \\
Gemma 4 31B & B & $-13.4$ & $-6.9$ & $-6.8$ & $-16.7$ \\
            & E & $-12.1$ & $-6.6$ & $-7.0$ & $-14.9$ \\
Qwen3.8 Flash & B & $-15.9$ & $-5.5$ & $-8.7$ & $-20.6$ \\
              & E & $-10.8$ & $-5.5$ & $-9.5$ & $-19.4$ \\
Fine-tuned & & $-9.8$ & $-8.0$ & $-9.5$ & $-12.1$ \\
\bottomrule
\end{tabular}
\caption{Robustness of the first-to-last quarter change in precision (percentage points) on scatter plots at $d=0.02$. One-to-one matching uses a Hungarian assignment that minimizes total distance, in which each ground-truth point can be matched to at most one prediction and matches farther than $d$ are discarded. Hidden removed drops fully hidden markers from the synthetic ground truth. One panel per article keeps the first panel of each article. All changes are significant ($p<0.01$, Wilcoxon signed-rank).}
\label{tab:app_robust}
\end{table*}

\end{document}